\documentclass[letterpaper, times, psfig, 10 pt, conference]{ieeeconf}
\IEEEoverridecommandlockouts
\usepackage[backend=biber,
            hyperref=true,
            url=false,
            isbn=false,
            doi=false,
            backref=false,
            style=ieee,
            natbib=true,
            citestyle=numeric-comp,
            sorting=none,
            block=none]{biblatex}
            
\renewcommand{\bibfont}{\small}
\usepackage{graphics}
\usepackage[pdftex]{graphicx}

\usepackage[font={small}]{caption}
\usepackage{subcaption}

\usepackage{mathtools}
\usepackage{amsmath, amssymb, amscd}
\usepackage{wasysym} 
\usepackage{nicefrac}       
\usepackage{mdwmath}
\usepackage{mdwtab}
\usepackage{eqparbox} 

\usepackage{array} 
\usepackage{tabularx}
\usepackage{multirow}
\usepackage{multicol}
\usepackage{booktabs}
\usepackage{colortbl}

\usepackage[T1]{fontenc} 
\usepackage[utf8]{inputenc}
\usepackage[english]{babel} 
\usepackage{units}
\usepackage{bm}
\usepackage{xspace}
\usepackage{flushend}
\usepackage{balance}

\usepackage{csquotes}
\usepackage{makeidx}

\let\labelindent\relax
\usepackage{enumitem}

\usepackage{soul} 
\usepackage{subfiles} 

\usepackage[yyyymmdd]{datetime}

\date{\protect\formatdate{1}{1}{2001}}

\usepackage{url}
\usepackage{color}
\usepackage[usenames,dvipsnames,table]{xcolor}

\makeatletter
\g@addto@macro{\UrlBreaks}{\UrlOrds}
\makeatother

\usepackage{marginnote}
\usepackage{soul} 

\newcommand{\tocite}[1]{%
\textcolor{red}{[cite:\ifthenelse{\equal{#1}{}}{}{#1}?]}
}

\newcommand{\ignore}[1]{}

\newcommand{\states}{\mathcal{S}}
\newcommand{\actions}{\mathcal{A}}

\newcommand{\rewards}{\ensuremath{\mathcal{R}}}
\newcommand{\observations}{\ensuremath{\mathcal{Y}}}

\begin{document}

\title{\Large \bf Visuomotor Mechanical Search: Learning to Retrieve Target Objects in Clutter}

\author{
Andrey Kurenkov$^{*1}$, %
Joseph Taglic$^{*1}$, %
Rohun Kulkarni$^{1}$, 
Marcus Dominguez-Kuhne$^{2}$, \\
Animesh Garg$^{3,4}$,
Roberto Mart{\'i}n-Mart{\'i}n$^{1}$, %
Silvio Savarese$^{1}$
\thanks{
$^{*}$These authors contributed equally.
}
\thanks{
\scriptsize{$^{1}$Stanford University, $^{2}$Caltech, $^{3}$University of Toronto \& Vector Institute $^{4}$Nvidia}
}
}



\maketitle

\begin{abstract}
When searching for objects in cluttered environments, it is often necessary to perform complex interactions in order to move occluding objects out of the way and fully reveal the object of interest and make it graspable. Due to the complexity of the physics involved and the lack of accurate models of the clutter, planning and controlling precise predefined interactions with accurate outcome is extremely hard, when not impossible. In problems where accurate (forward) models are lacking, Deep Reinforcement Learning (RL) has shown to be a viable solution to map observations (e.g. images) to good interactions in the form of close-loop visuomotor policies. However, Deep RL is sample inefficient and fails when applied directly to the problem of unoccluding objects based on images. In this work we present a novel Deep RL procedure that combines i) teacher-aided exploration, ii) a critic with privileged information, and iii) mid-level representations, resulting in sample efficient and effective learning for the problem of uncovering a target object occluded by a heap of unknown objects. Our experiments show that our approach trains faster and converges to more efficient uncovering solutions than baselines and ablations, and that our uncovering policies lead to an average improvement in the graspability of the target object, facilitating downstream retrieval applications.
\end{abstract}

\IEEEpeerreviewmaketitle

\section{Introduction}
\label{sec:intro}

It is challenging for a robot to retrieve a known target object from a pile of cluttered elements, even when the target is partially visible. The occluding objects make it or impossible to grasp the desired object, requiring the robot to interact first with the unknown clutter to improve the target's \textit{graspability}. Such situations appear frequently in domains such as home robotics or even logistic centers, and is considered an instance of the Mechanical Search problem~\cite{danielczuk2019mechanical}. 

Previous approaches proposed carefully-coded heuristics~\cite{danielczuk2019mechanical,hermans2012guided} or learned~\cite{yang2020deep,zeng2018learning} sequences of actions that try to discover and retrieve the desired object.
In both cases, the problem is simplified by choosing the action space to be a set of linear pushes parameterized as a point on the clutter and a direction to push, and a retracting motion after each action. The simplified pushing strategy leads to longer execution times and undesired clutter motion due to the retraction motion. Further, the goal of the pushing motions is sometimes primarily to singulate the target object, not to uncover it from underneath the covering clutter. A more natural solution for uncovering is to use a closed-loop and continuous pushing strategy based on the current visual signal, allowing the policy to adapt to the unforeseeable reactions of the interactions with the clutter.

\begin{figure}[!t]
\centering
\vspace{-5pt}
    \includegraphics[width=\linewidth]{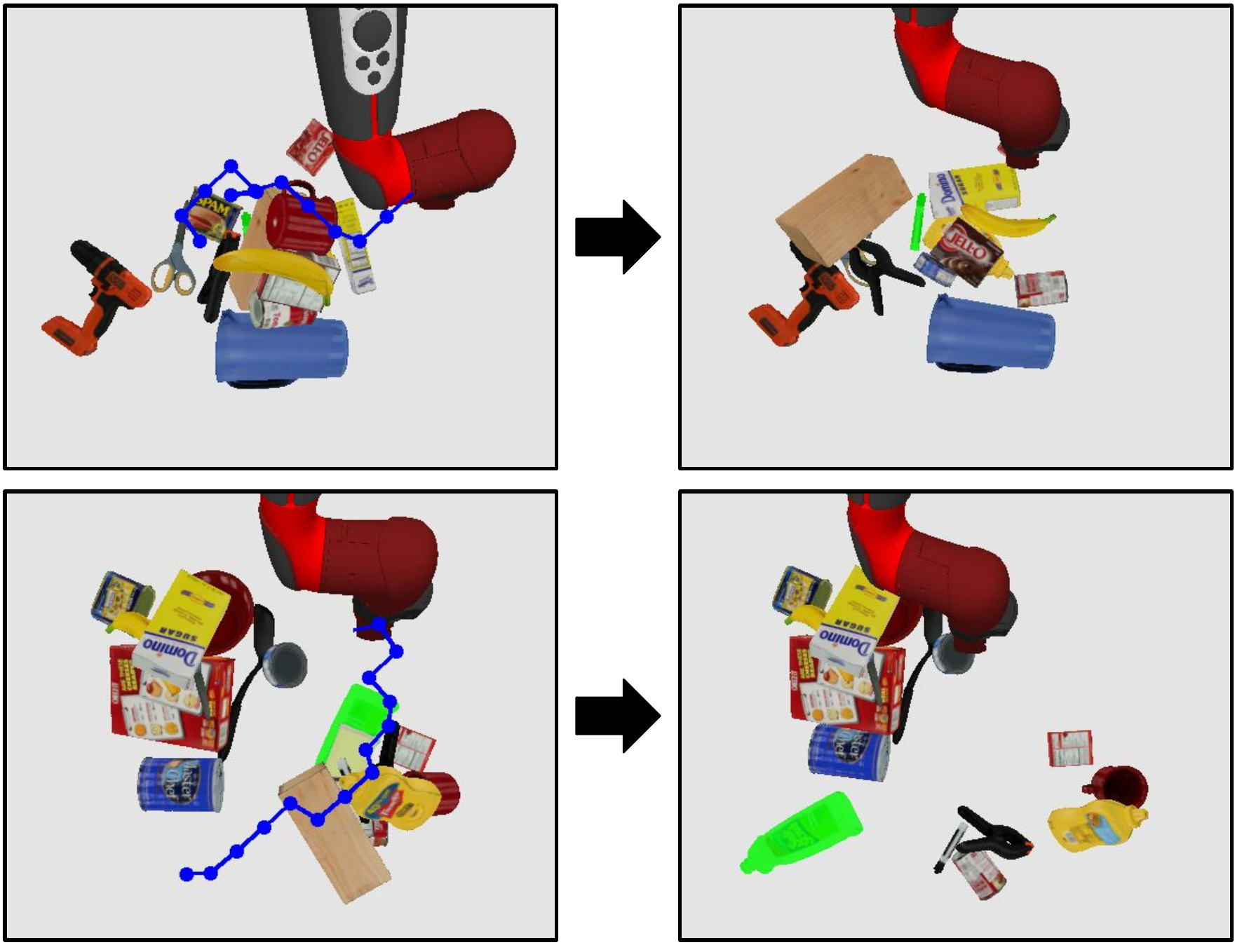}
    \caption{An overview of the problem we address. On the left, the start state with the visible part of the target object highlighted green and the trajectory our policy took shown in blue. On the right, the resulting final state, with the target object now more visible and graspable.}
  \label{fig:intro-fig}
  \vspace{-15pt}
\end{figure}

In this work, we propose to address the problem of uncovering a partially visible target object to improve graspability by learning a visuomotor policy that maps the current image of the cluttered pile to continuous robot actions. 
We propose to use deep Reinforcement Learning (RL) to learn such a policy, given the recent successes of deep RL in image-based sequential decision making problems with unknown or complex environment dynamics~\cite{kalashnikov2018qt,xiazamirhe2018gibsonenv,tan2018sim}. However, existing RL approaches are data hungry and brittle: they require a large number of environment interactions to learn a mapping from high dimensional images to successful continuous robot actions, and often the algorithms fail to find a solution. A common strategy to avoid having to collect many interactions in the real world, which can be slow and dangerous, is to use a simulator. Nevertheless, for hard interactive problems, existing RL algorithms for continuous control may still struggle to learn a successful strategy in simulation.
Therefore, in this work we present a deep RL solution combining in a novel manner three algorithmic strategies that allow our method to learn to uncover the target object based on images.

A first strategy to improve the efficiency of visuomotor learning in simulation is to leverage the information about the state of the environment from the simulator. Such \textit{privileged information} is used only during training, while the component that maps inputs to actions at test time (the actor) is trained to use only images. While promising, this strategy alone is not sufficient to learn complex multi-object continuous tasks such as pushing to uncover a target because of the large state space for exploration~\cite{pinto2017asymmetric} 

A second algorithmic strategy to improve RL training is to guide the exploration using teachers~\cite{kurenkov2019ac}. Teachers are expert policies that provide suggestions for suboptimal actions, which can guide the exploration of the RL agent to the relevant areas of the state space. While this helps with exploration, the challenge of learning a policy directly from images that can be used on a real robot is still significant.

For this last challenge, a third algorithmic strategy that has been shown demonstrated is to provide the agent with inputs in a mid-level representation instead of directly the raw RGB pixel inputs. Learning in the mid-level representation is more effective and facilitates transfer from simulation onto a real robot~\cite{xiazamirhe2018gibsonenv}. We make use of this concept by leveraging the segmentation mask of the target object, similarly to \cite{yang2020deep, kim2020acceleration}, and the known extrinsics and intrinsics of the camera to get the position of these pixels relative to the end effector and provide those as input to the agent. 

In summary, the main contributions in this work are:
\begin{enumerate}[leftmargin=*]
\item a novel learning procedure that combines an asymmetric architecture to leverage privileged information, guidance from suboptimal teachers, and mid-level representations to train deep RL agents for visuomotor continuous control tasks,
\item the application and instantiation of this learning procedure to solve the problem of uncovering a target object to improve its graspability.
\end{enumerate}

We conducted extensive experiments in simulation to evaluate the performance of our learned agents. The results indicate that our combination of privileged information, teacher guidance, and mid-level representation greatly improves sample efficiency and final performance. The method, applied to our Mechanical Search problem, learns to uncover a target object under clutter and improves its graspability.

\section{Related Work}
\label{sec:rw}

When operating in unstructured environments, robots often encounter cluttered environments that need to be interacted with. This may occur, for example, during sorting or retrieving a specific object. This problem has been tackled from different perspectives. Interactive perception approaches~\cite{bohg2017interactive} have considered the perceptual problem of segmenting images of a pile of objects into coherent components that move together~\cite{martin2016integrated,van2012maximally,kenney2009interactive,katz2014perceiving}. Several works in this area applied pushing actions to facilitate segmentation~\cite{hermans2012guided,gupta2014using,eitel2020learning} but not to uncover a target object. 

The grasping community considered the problem of planning and executing grasps on a pile of cluttered objects until all objects have been cleared. Depending on the assumed prior knowledge these methods can be considered model-based~\cite{berenson2008grasp,moll2017randomized,xiao2019online} or model-free approaches~\cite{srivastava2014combined, mahler2017learning, danielczuk2019mechanical, novkovic2019object, yang2020deep, berscheid2019robot, jang2017end,zeng2018learning}. The latter use only images to decide on the best grasping action for a pile of objects. A recent work~\cite{roboturk-real} presented an instance of this problem to show how human demonstrations for such a task can be crowdsourced. Differently, our goal is not to clear a pile of objects, but to uncover and facilitate grasping of a given target object.

The two works that are the closest to ours are~\cite{danielczuk2019mechanical, yang2020deep}. Danielczuk et al.~\cite{danielczuk2019mechanical} defined the problem of Mechanical Search, searching for a known object among unknown cluttering objects with interactions, and proposed a method that chooses among discrete pre-specified action policies (e.g. pushing, grasping with suction, grasping with parallel-jaw gripper) based on the acquired RGB-D images. 
While their method includes a heuristically pre-specified action to push clutter, we go further in this work and use RL to optimize continuous visuomotor policies to more optimally push objects in the environment to uncover a known target. We consider our method complementary to the one by Danielczuk et al.~\cite{danielczuk2019mechanical}: our learned pushing policies could be integrated into their framework as a more robust action policy.

Yang et al.~\cite{yang2020deep} proposed a Bayesian exploration policy to search for a target object with pushing actions. They pose the action selection problem as a Q-learning problem in the image domain. 
In contrast to their approach (and also to the one by Danielczuk et al.~\cite{danielczuk2019mechanical}), we aim to learn continuous controlled actions without retracting the arm after each discrete push, so that the manipulation is more reactive to the outcome of the interaction.

\begin{figure*}[!ht]
  \centering
      \begin{subfigure}{0.45\textwidth}
        \centering
        \includegraphics[width=\linewidth]{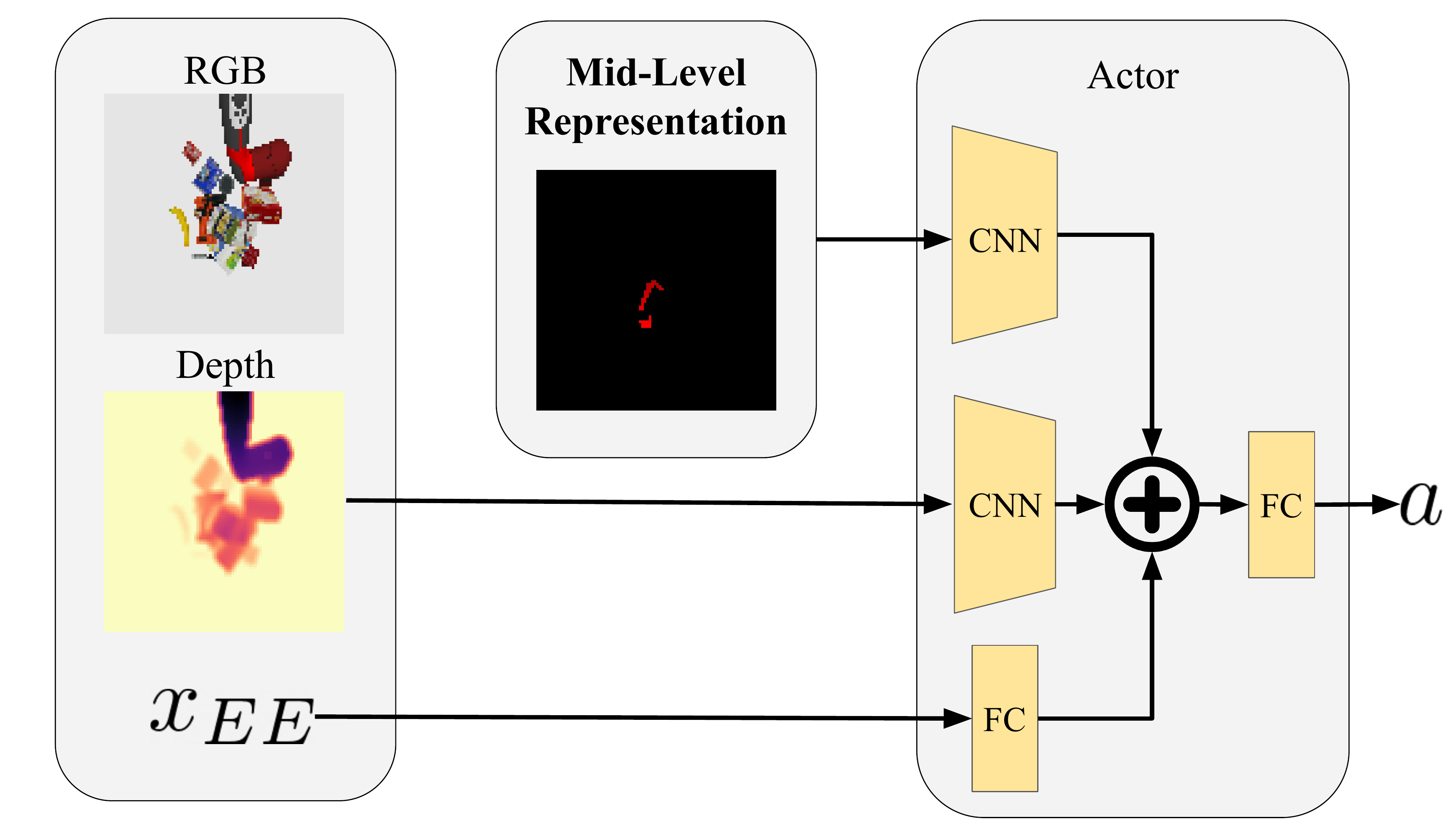}
        \caption{Actor of our actor-critic architecture using our mid-level representation. Inputs to our algorithm are RGB-D images, a segmentation image of the target object, and the pose of robot's end-effector. We transform the input into a mid-level representation consisting of a position image (each pixel has the values of the 3D offset to it from the end effector, similarly to \cite{liu2018intriguing} pixel coordinate channel) masked by the segmentation of the target object as in \cite{yang2020deep,kim2020acceleration}. The image in this representation, the depth image and the end-effector pose are featurized by separate network heads, concatenated and used to generate a pusing action.}
    \end{subfigure}%
    \hfill
    \begin{subfigure}{0.5\textwidth}
        \centering
        \includegraphics[width=\linewidth]{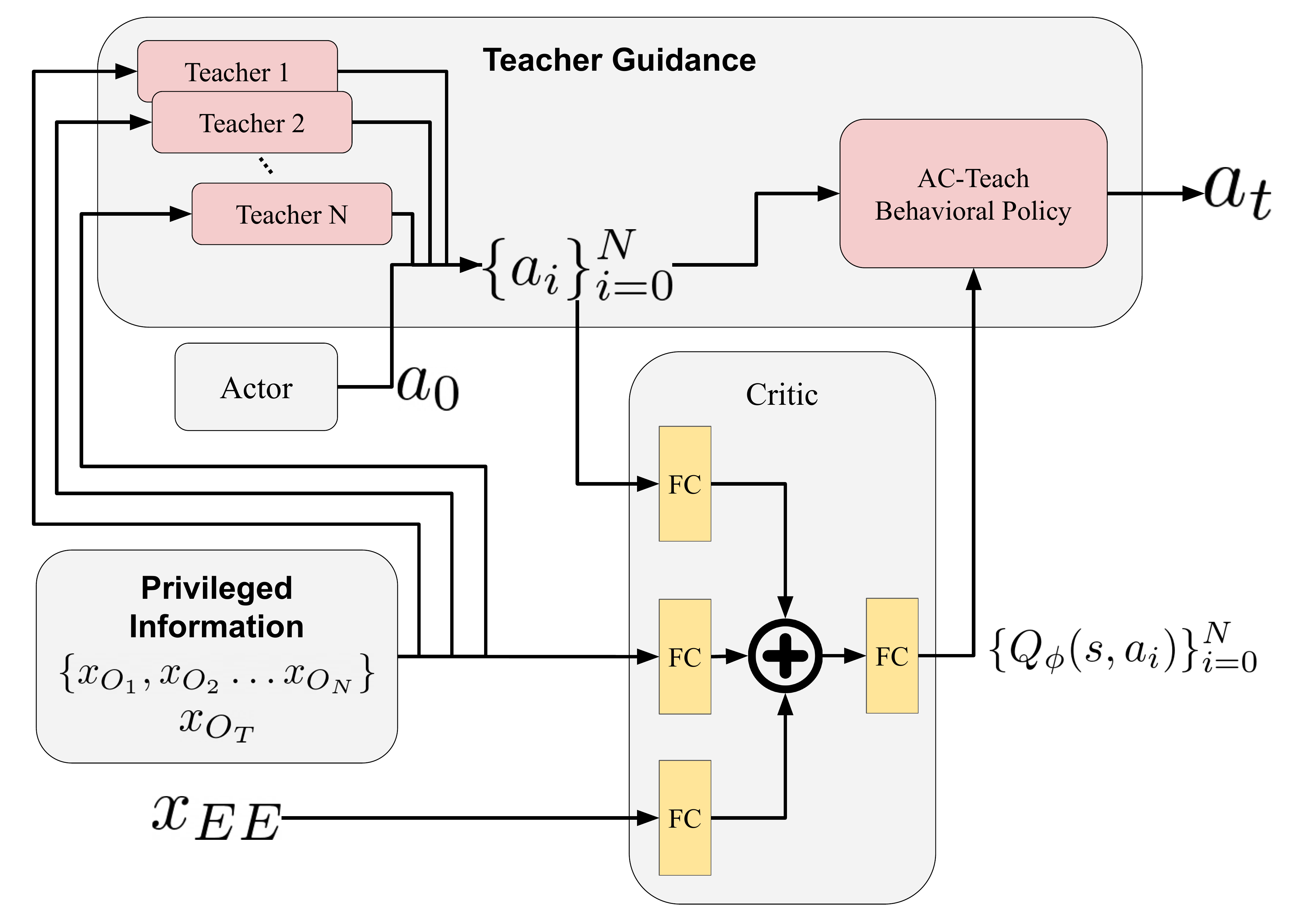}
        \caption{Training architecture leveraging teacher guidance and privileged information. Both the teachers and the critic of our actor-critic RL architecture leverage privileged information: the pose of all objects in the cluttered pile, including the target. The different teachers provide strategic pushing motions (straight, spiral, \ldots). The teachers and the critic with privileged information act only during training to reduce the samples necessary to train the actor.}
    \end{subfigure}%
  \caption{An overview of our method. $x_{EE}$ denotes the end effect position of the robot, $x_T$ denotes the position of the target object, $\{x_{O_1}, x_{O_2} \dots x_{O_N}\}$ denotes that set of the non target objects, and actions in the form of end effector position control commands are denoted with $a_i$. Our method is the result of combining the actor policy in (a) and the training procedure in (b).}
  \vspace{-10pt}
  \label{fig:method}
\end{figure*}


Learning continuous interaction with a complex environment using RL is a hard exploration problem. The exploration can be simplified using teachers, black box expert policies that can be queried during training time. This paradigm has been shown successful for manipulation~\cite{kurenkov2019ac}; in this work we show, for the first time, its applicability to high-dimensional inputs, i.e. images. 

Although teacher guidance provides a better exploration strategy for the task, it does not address the additional challenge of learning from images. This can be alleviated using privileged information during training. We use privileged in our teachers an in the critic of our actor-critic solution, similar to Pinto et al.~\cite{pinto2017asymmetric}. Additionally, recent work has studied how to use pretrained representations for robot learning~\cite{lee2019making,raffin2019decoupling,sax2018mid}, and in particular \cite{yang2020deep} showed how to use a pre-trained segmentation module to direct the policy towards the target object. We take inspiration from these methods and propose the use of a position image (similar to \cite{liu2018intriguing} use of a pixel coordinate channel), masked with the segmentation image, as an intermediate representation to facilitate policy learning.

\section{Method}
\label{sec:ps}

In the following, we first explain how we pose the problem of uncovering a known object with continuous pushing interactions based on visual images as a reinforcement learning problem. Then, we explain the novel policy architecture we proposed to solve this problem, followed by the special training procedure that leads us to a trained policy in simulation that we can transfer to the real robot.

\subsection{Problem Statement}


In our setup, a stationary robot is tasked with uncovering a known object of interest from the unknown occluding objects on top using pushing actions based on the images acquired by an RGB-D sensor. 
We pose this problem as a reinforcement learning problem on a Markov Decision Process (MDP) defined by the tuple $(\states, \actions, \rewards, \observations, \gamma, \rho)$. In the tuple, $\states$ is a continuous state space, $\actions$ is a continuous action space, $\observations$ is the space of observations, $\rewards$ is a reward function $\mathcal{R}(s,a) = r \in \mathbb{R}$, $\gamma \in [0,1)$ is a discount factor (for infinite horizon problems) and $\rho$ is the initial state distribution. The goal is to learn a policy $\pi(a|s) = p(a|s)$ that selects actions based on current observations so as to maximize the expected reward~\cite{sutton2018reinforcement}.

The instantiation of our problem is depicted in Fig.~\ref{fig:method}. In our problem, the observations are RGB-D images and the position of the robot's end-effector in Cartesian space, and the actions of the agent are small end-effector changes in position (offsets relative to the current position). 

\noindent \textbf{Reward Function}. The reward function that represents our task provides positive feedback when the object of interest becomes less occluded, and negative feedback when objects are moved. These penalties deter the agent from learning to push the entire pile to spread all objects, a strategy that can be dangerous and break the objects or the robot in the real world. Concretely, the reward function is a sum of the following terms:
\begin{enumerate}[leftmargin=*]
    \item \textbf{Target Uncovering Reward} of $2.5 * c$, where $c$ is change in object visibility computed as $c = (occlusion_{t-1}$ $- occlusion_t)/occlusion_{t-1}$. The target object to uncover is not static, but is rather a function of the state and changes every episode.
    
    \item \textbf{Heap Movement Penalty} If $c < 0.05$, we penalize moving other objects needlessly. For each object, if its change in position is $m_o =  \|(pos_t - pos_{t-1})\|$ , this term is $75.0/N * m_o$. $m_o$ is represented in meters, and therefore typically $m_o < 0.05$.
    
    \item  \textbf{Target Movement Penalty} If $m_t$ is change in object visibility computed as $m_t = \|(target\_pos_t  - target\_pos_{t-1} )\|$, 75.0 * $m_t$. $m_t$ is represented in meters, and therefore typically $m_t < 0.05$.
    
    \item  \textbf{Workspace Limits Penalty} A -$0.5$ reward is provided to encourage the agent to avoid the bounds of the workspace.
    
    \item  \textbf{Idleness Penalty} A -$0.5$ reward is provided to encourage the agent to seek positive rewards despite of the risk of other negative rewards.
\end{enumerate}

An episode is considered finished when the target object is completely visible, or a specified limit of actions is reached.

\subsection{Closed-Loop Visuomotor Control Policy}
In addition to the complexity of the above reward function, this RL problem is challenging due to the difficulty of having to adapt to uncovering different target objects, due to having to explore a large state space, and due to having to learn from high dimensional visual inputs. We design our policy and training algorithm with a novel combination of features designed to address these challenges.

Though we could train our policy directly from the RGB-D observations of the scene, this would present the policy with the challenge of having to detect every kind of object it may have to uncover just from the signal of rewards. Therefore, instead of having the agent learn from RGB images, we provide it with a mid-level representation that encodes the approximate position of the target object.  

To make this representation, we first mask the pixels of the depth image to only those belonging to the target object, as in \cite{yang2020deep, kim2020acceleration}. Then, we use the known camera intrinsics and extrinsics to back-project the pixels into their 3D positions relative to the position of the robot's end effector. We base this second step on previous work that showed that CNNs are not well suited regress the coordinates of non-zero pixels in an image (something the agent will need to do to move towards the target object), and that explicitly providing the position of these pixels in the input is an effective solution \cite{liu2018intriguing}. The result is a 3 channel 'image', in which the 3 channels of the non zero pixels are the X, Y, and Z offsets from the end effector to visible pixels of the target object in meters. 

For implementing this mid-level representation, we use the ground truth segmentation mask from rendering in the simulator. For eventual transfer to real world settings, it should also be possible to instead use a model trained for segmenting the objects of interest, such as the encoder-decoder segmentation network from previous work~\cite{wang2019densefusion}. 


The deterministic actor $\pi_{\theta}(s)$ then takes this mid-level representation as input, as well as a depth image of the tabletop and heap of objects. Additionally, the actor accepts the position of the end effector as an input. Each image input is separately processed by a 3-layer convolutional neural net with the same structure as is used in~\cite{lillicrap2015continuous}. The end effector position is passed through a fully connected layer of size 32 and ReLU activations. The outputs of the last convolution layers are then flattened and concatenated with each other and the end effector position output, and these features are processed by two fully connected layers of size 256. Lastly, a final fully connected layer with a tanh activation produces the action output scaled to the appropriate range.

\begin{figure*}[!th]
\centering
\begin{subfigure}{0.95\textwidth}
\centering
\includegraphics[width=\linewidth]{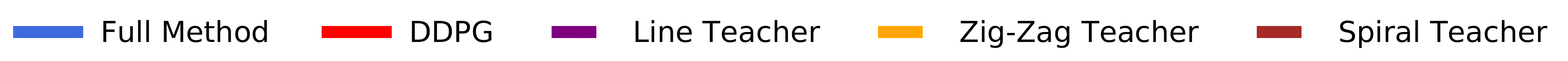}
\end{subfigure}%
\\
\begin{subfigure}{0.24\textwidth}
\centering
\includegraphics[width=\linewidth]{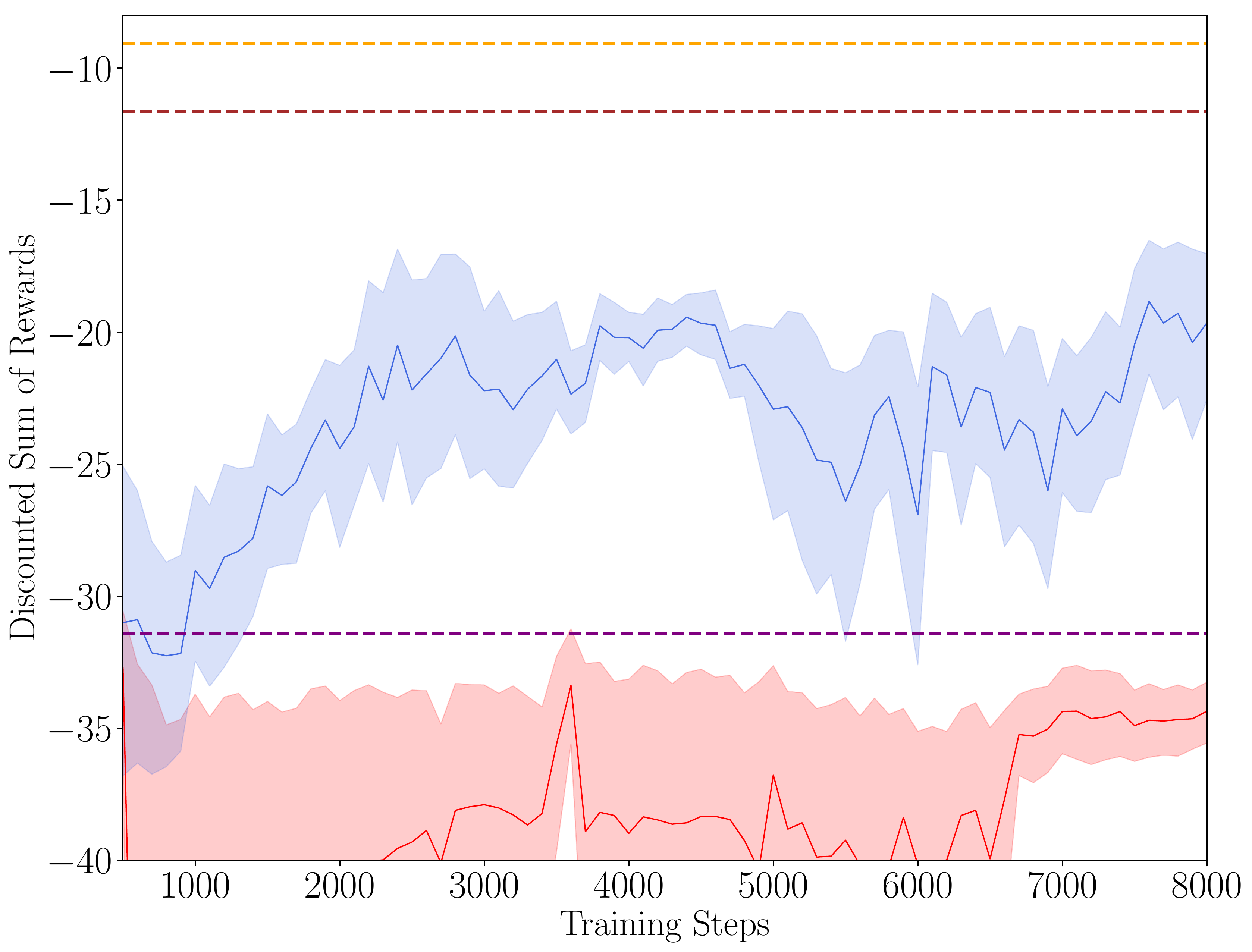}
\end{subfigure}%
\hfill
\begin{subfigure}{0.24\textwidth}
\centering
\includegraphics[width=\linewidth]{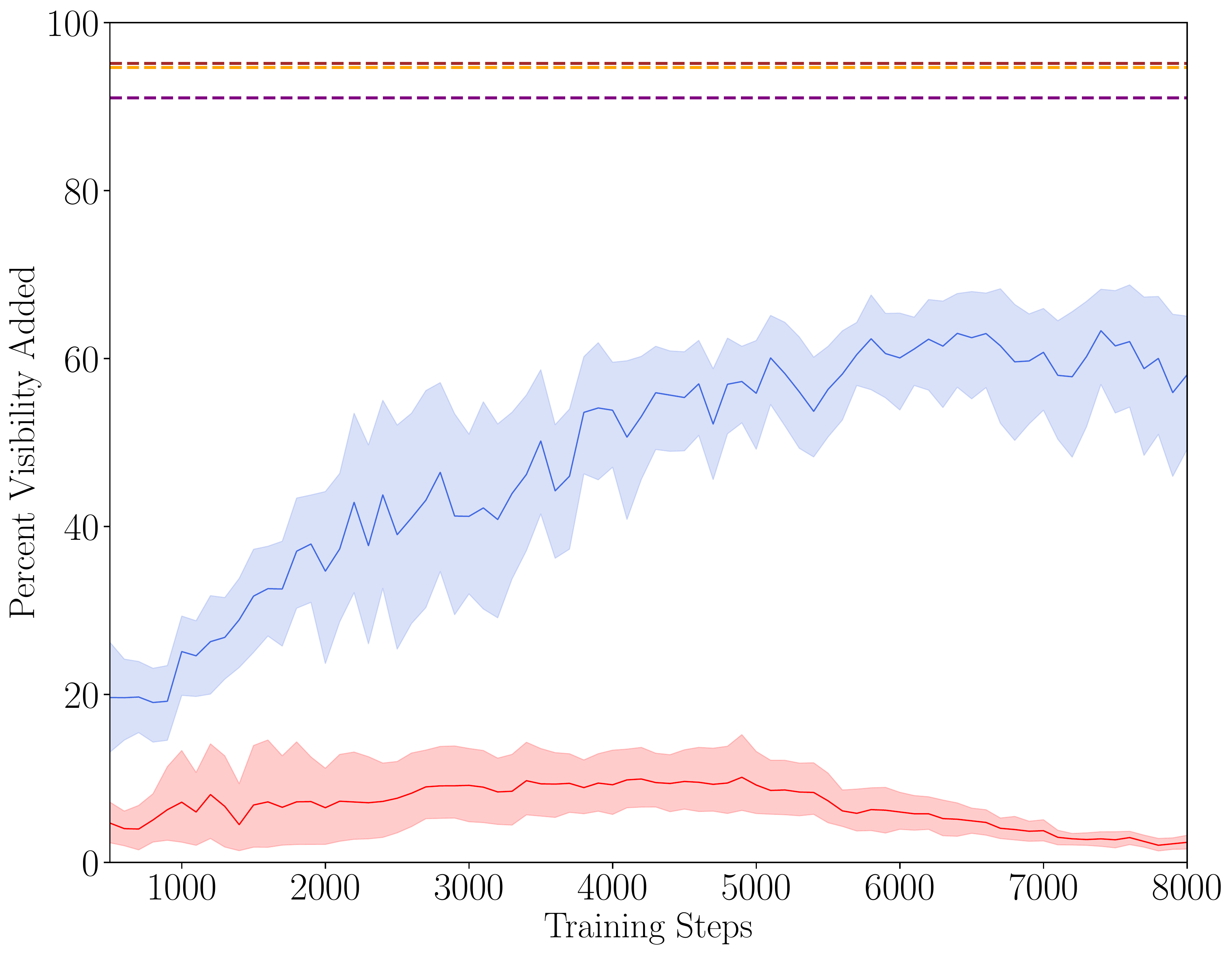}
\end{subfigure}%
\hfill
\begin{subfigure}{0.24\textwidth}
\centering
\includegraphics[width=\linewidth]{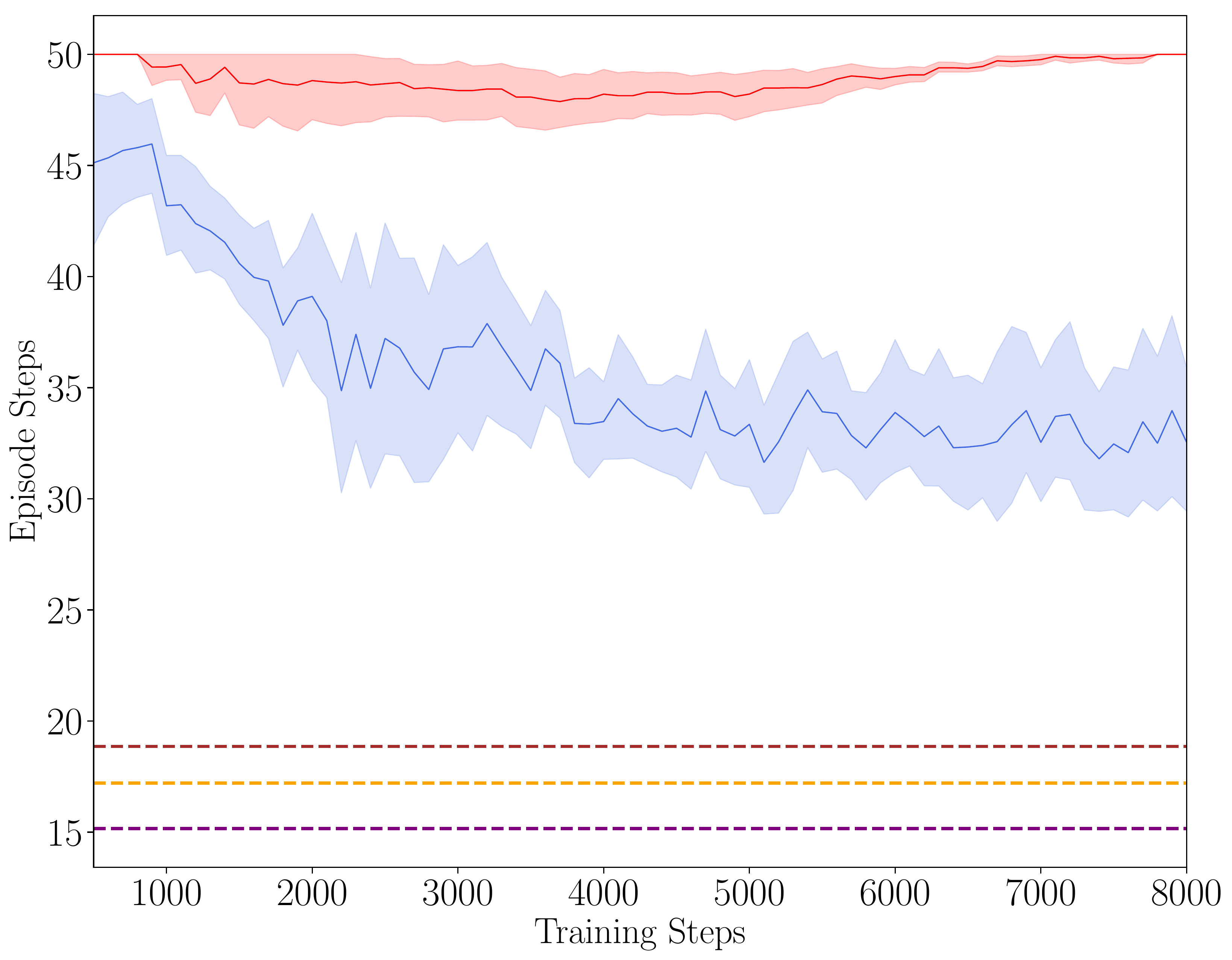}
\end{subfigure}%
\hfill
\begin{subfigure}{0.24\textwidth}
\centering
\includegraphics[width=\linewidth]{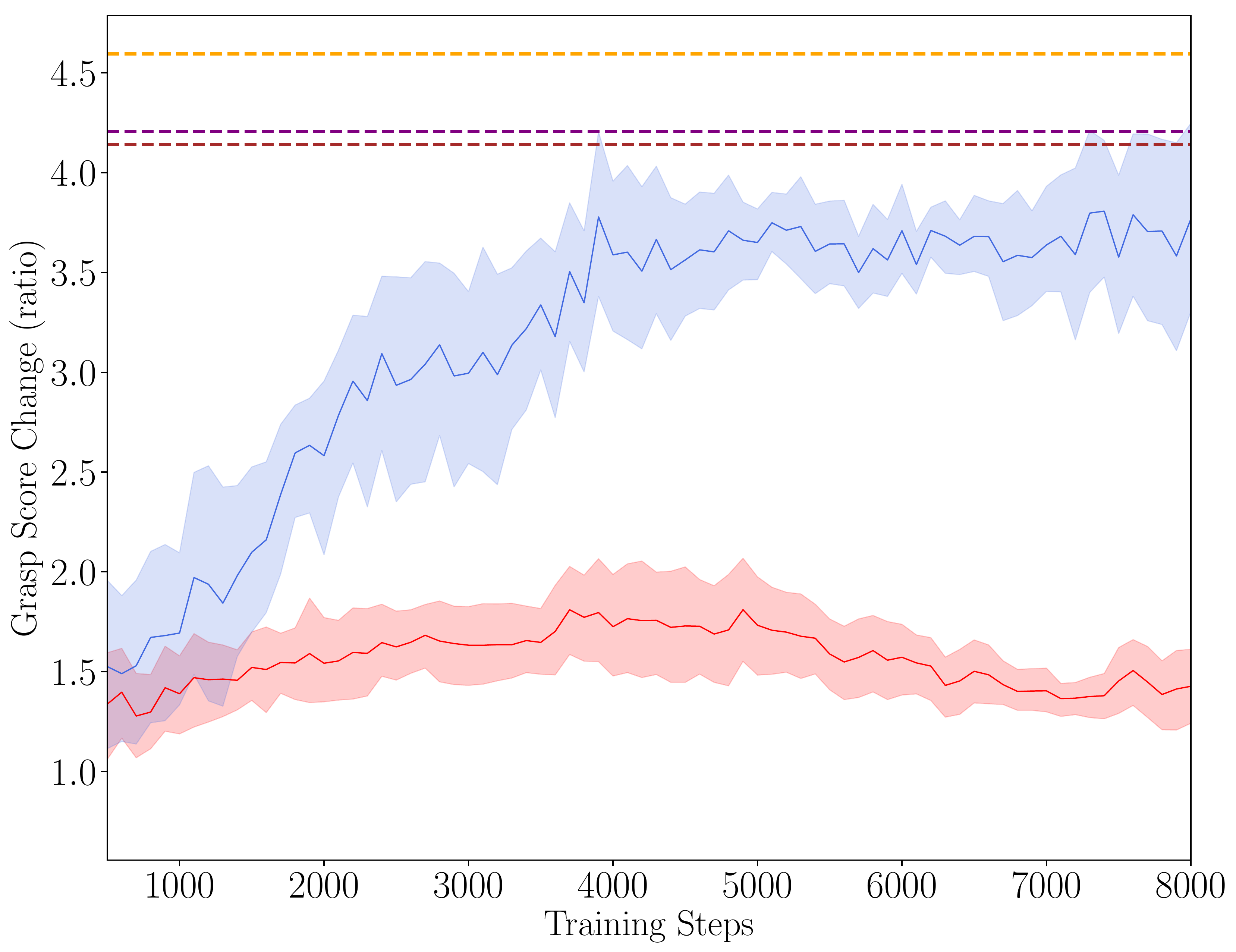}
\end{subfigure}%
\\
\begin{subfigure}{0.24\textwidth}
\centering
\includegraphics[width=\linewidth]{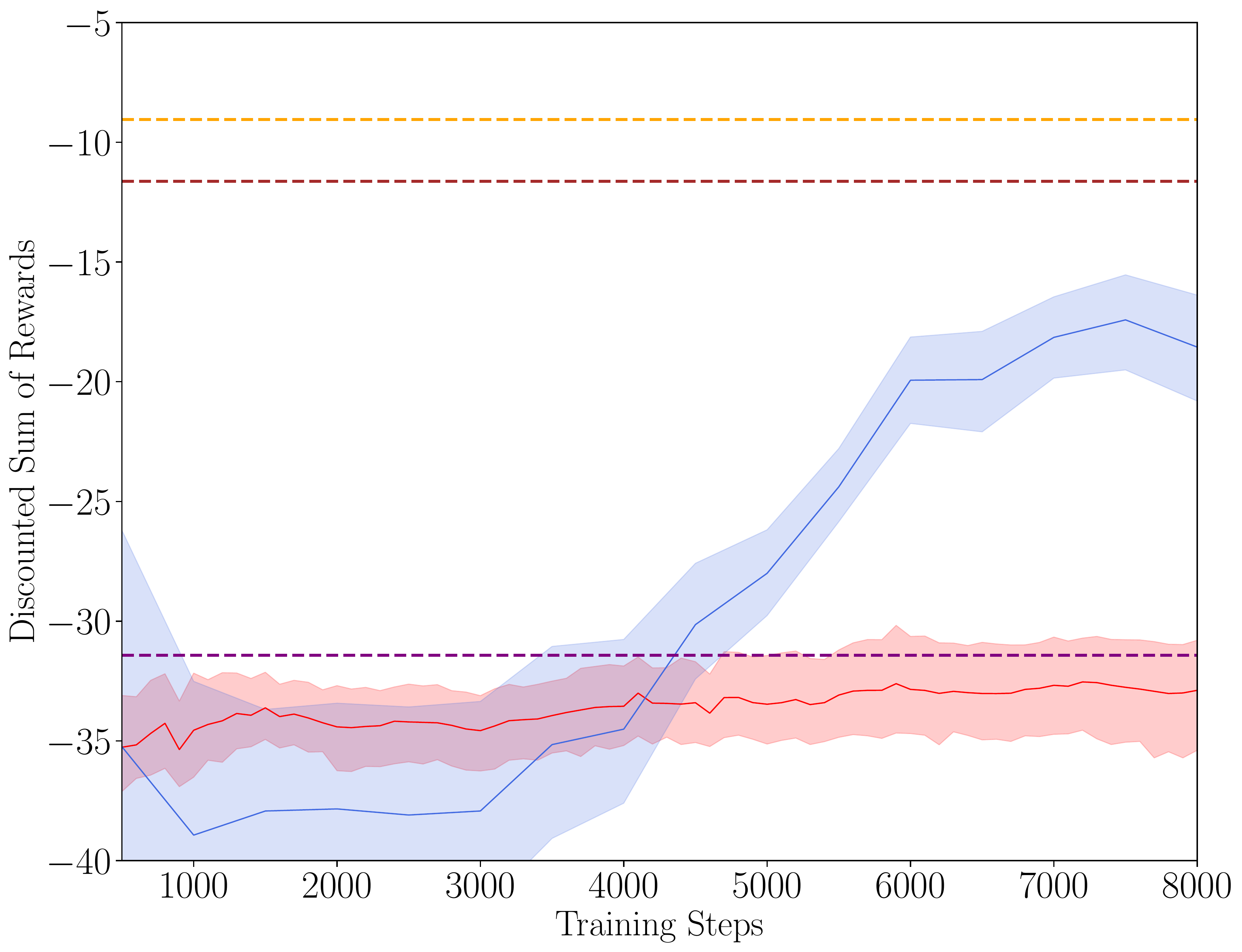}
\caption{}
\end{subfigure}%
\hfill
\begin{subfigure}{0.24\textwidth}
\centering
\includegraphics[width=\linewidth]{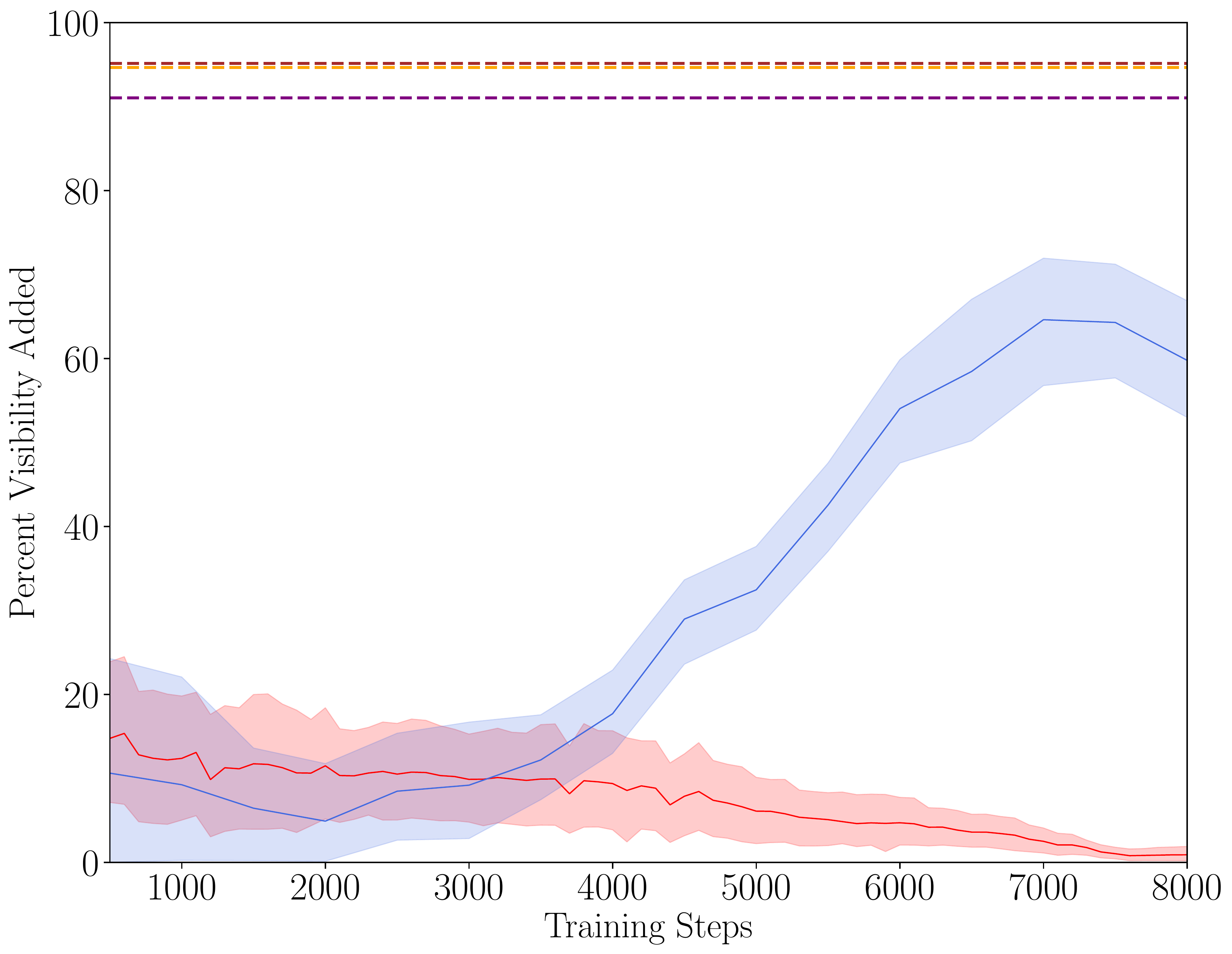}
\caption{}
\end{subfigure}%
\hfill
\begin{subfigure}{0.24\textwidth}
\centering
\includegraphics[width=\linewidth]{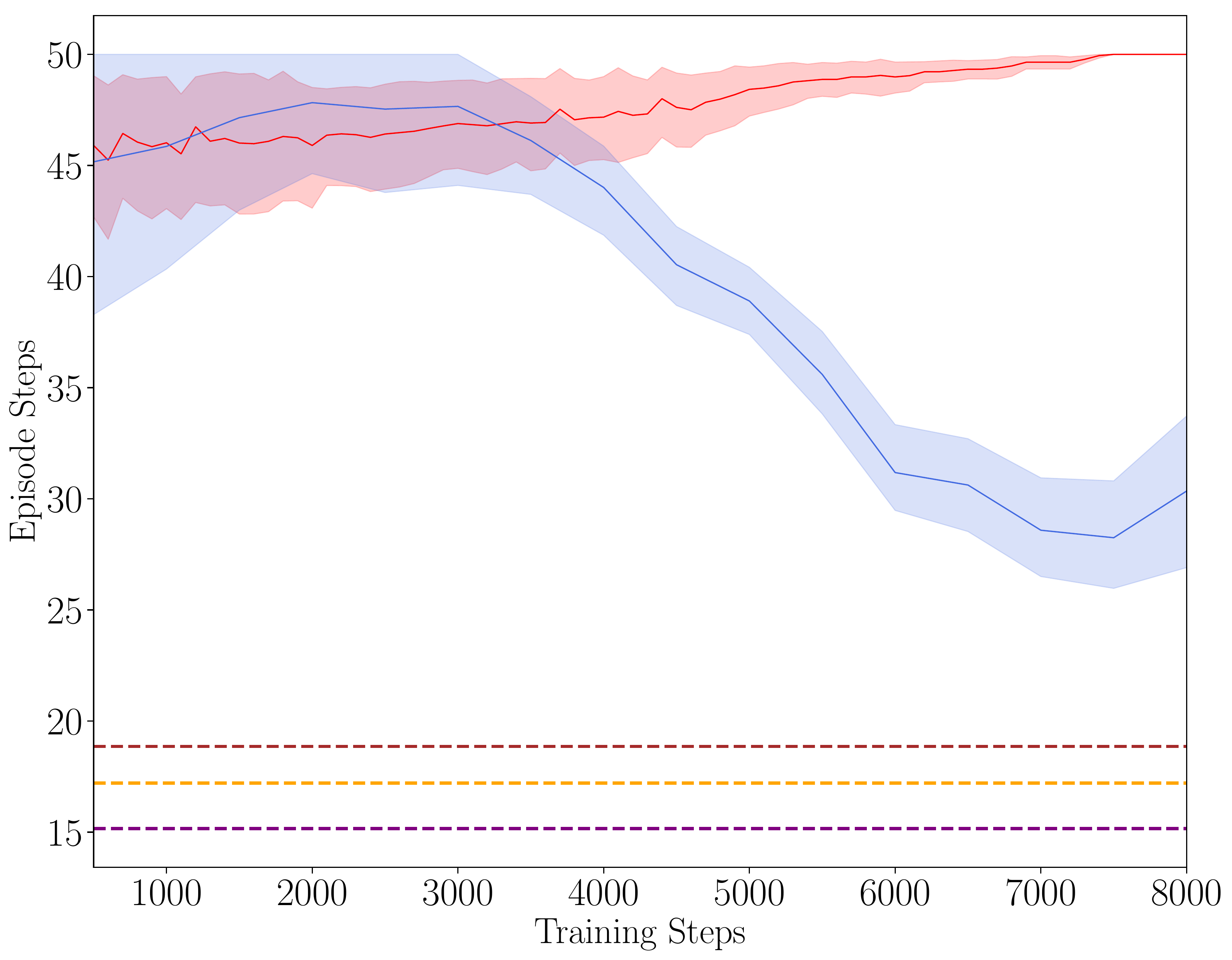}
\caption{}
\end{subfigure}%
\hfill
\begin{subfigure}{0.24\textwidth}
\centering
\includegraphics[width=\linewidth]{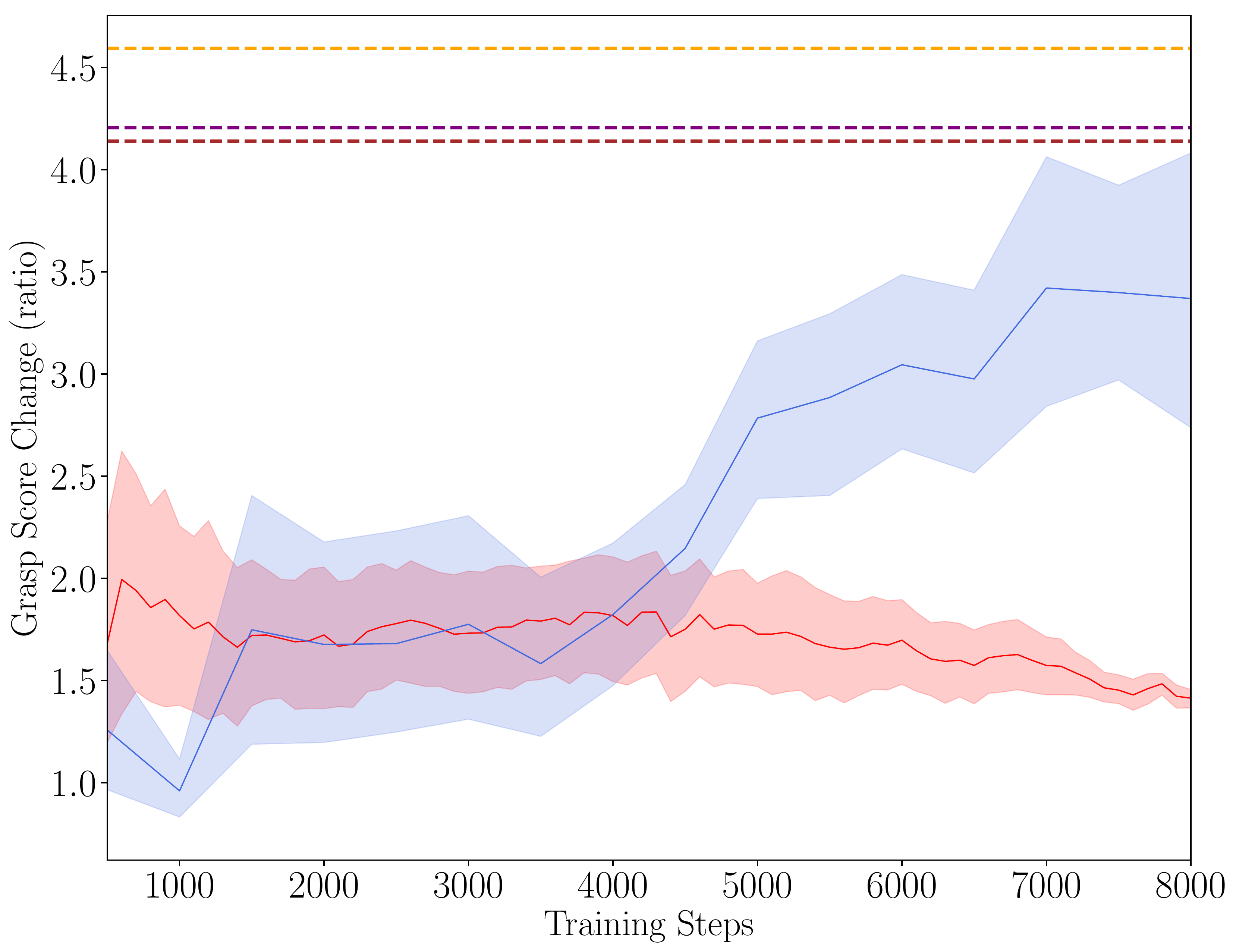}
\caption{}
\end{subfigure}%
\hfill

\caption{Quantitative results in simulation for experiments with single-heap (top) and dual-heap (bottom) conditions showing evaluation results throughout training for the episode rewards (a), change in target visibility (b), number of steps taken until the episode finished (c), and change in target object graspability (d) respectively. Each plot represents values attained by the agent without added noise on held out evaluation heaps, plotted with respect to number of actions taken in the environment as part of exploration for training. The plots are limited to end on the step when the trained agent reaches its peak performance. Teacher evaluations are averages evaluated based on 100 rollouts. As shown by columns (a)-(c), our method consistently learns to improve these metrics and gets close to matching the teachers. And as shown in column (d), our method attains improvement in graspability that are close to on par with the teachers, despite not having access to the priviliged information they rely on. }
\label{fig:sim-results}
  \vspace{-15pt}
\end{figure*}

\subsection{Teacher-Guided Actor-Critic Policy Learning}
By its nature as a continuous control problem, our task requires many actions to be executed to achieve the goal, which makes effective exploration challenging. For this challenge, we adopt the idea of agent exploration being guided by a set of provided black box policies (teachers) that suboptimally address part of the task and can suggest possible actions to take in any state. Early on in training the teachers' action suggestions are still expected to be superior to exploration only guided by noise added to the actor's output, which enables the agent to train faster while also optimizing for disturbing the heap of objects less than the teachers. Pushing is a particularly good fit for this approach, since it is easy to come up with several heuristic solutions that would be expected to be suboptimal but may be better than random.  Specifically, we utilize the following teachers: 

\textbf{Straight Line Push:} Executes a random straight line push to execute above the target object.

\textbf{Zig Zag Push:} Same as above, but the end effector moves in a zig-zag pattern while moving along the straight line.

\textbf{Spiral Push:} The end effector is placed near the target object, and the arm spirals out from that location for a specified amount of distance.

To have access to the target object's location as is necessary to implement these teachers, and to be able to compute the task's reward function, we perform training in the PyBullet simulator~\cite{coumans2017bullet}, with the renderer from the Gibson v2 simulator~\cite{xiagibson2019}. 

Because of our setting of continuous control and having access to multiple teachers, we base our approach on that of~\cite{kurenkov2019ac}. Thus, as in that work the agent to be trained is based on a probabilistic variant of the Deep Deterministic Policy Gradient (DDPG) algorithm~\cite{henderson2017bayesian}, though other actor-critic algorithms could in principle be used. We make the critic probabilistic so that it can be used to select between the policy's chosen action or the actions output by hand-coded suboptimal teachers to the task for any given state. For more details including the exact formulations of the losses, refer to~\cite{kurenkov2019ac}.

While teacher guidance helps lessen the exploration challenge, it does not alleviate the challenge of learning from high dimensional image inputs, and so is not enough to enable our method to learn effectively and efficiently. To deal with this last challenge, we again make use of the 'privileged information' afforded to us by training in a simulator by making the critic only depend on this low-dimensional information for its input, as in~\cite{pinto2017asymmetric}. This privileged information is made up of two vectors, the position of the target object and a concatenation of the positions of all the other objects in the environment. All positions are provided relative to the end effector's position.

Both privileged information vectors and the end effector position are each processed with a separate fully connected network of size 32 and ReLU activations -- unlike the actor network, there is no need to train convolutional neural nets for the critic. The outputs of these layers are concatenated, and then processed with two fully connected layers of size 256 and ReLU activations and a final fully connected layer that outputs the Q value estimate. 

To summarize, on a given training step the policy's actor makes use of depth inputs and a mid-level representation to output its action, and our set of teachers also each output their respective actions. Then, the critic makes use of privileged information to evaluate each action so that the best one may be selected. Once an action is executed, its transition $(s,a,s',r)$ is put in a replay buffer. Transitions from the replay buffer are intermittently sampled as training data for the actor and critic. The critic is trained via the Bellman residual loss $\mathcal{L}_{\text{critic}} = (r + \gamma Q_{\phi'}(s', \pi_{\theta'}(s')) - Q_{\phi}(s, a))^2$ and the actor is trained with a deterministic policy gradient update to choose actions that maximize the critic $\mathcal{L}_{\text{actor}} = -Q_{\phi}(s, \pi_{\theta}(s))$ where $\phi'$ and $\theta'$ denote the use of target critic and actor networks. 
\section{Experiments}
\label{s_expp}
In our experiments, we aim to answer two main questions: First, we evaluate whether the proposed combination of asymmetric, teacher-guided visuomotor learning with a mid-level representation achieves better sample efficiency and task performance in terms of rewards than an approach not utilizing these ideas. And second, we evaluate if the proposed approach is beneficial for continuous visuomotor control to uncover known target objects and increase their graspability. Concretely, we will answer the following questions:
\begin{enumerate}
\item Does our method results in policies that consistently uncover the object and increase the expected graspability?
\item How quickly objects are uncovered, how much are objects moved as a result, and what is the trade off between time to uncover the object and amount by which objects are moved?
\end{enumerate}

To answer these questions, we perform a series of tests in simulation, with comparisons to several ablation versions of our solution.

\begin{figure*}[!t]

\centering
\includegraphics[width=\linewidth]{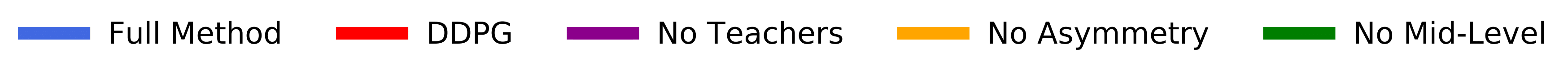}
\centering
\begin{subfigure}{0.45\linewidth}
  \includegraphics[width=\linewidth]{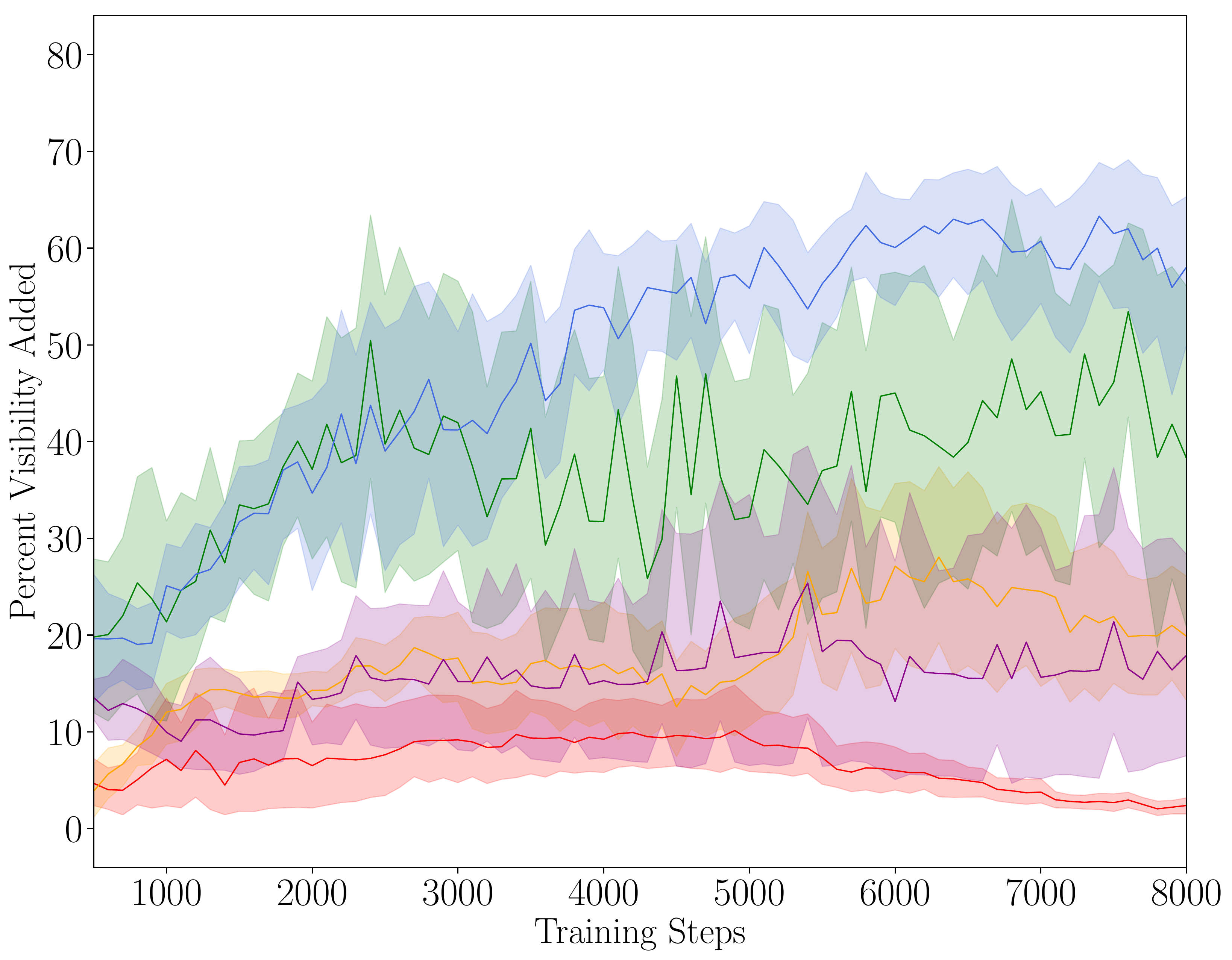}
  \end{subfigure}
  \hfill
\begin{subfigure}{0.45\linewidth}
  \includegraphics[width=\linewidth]{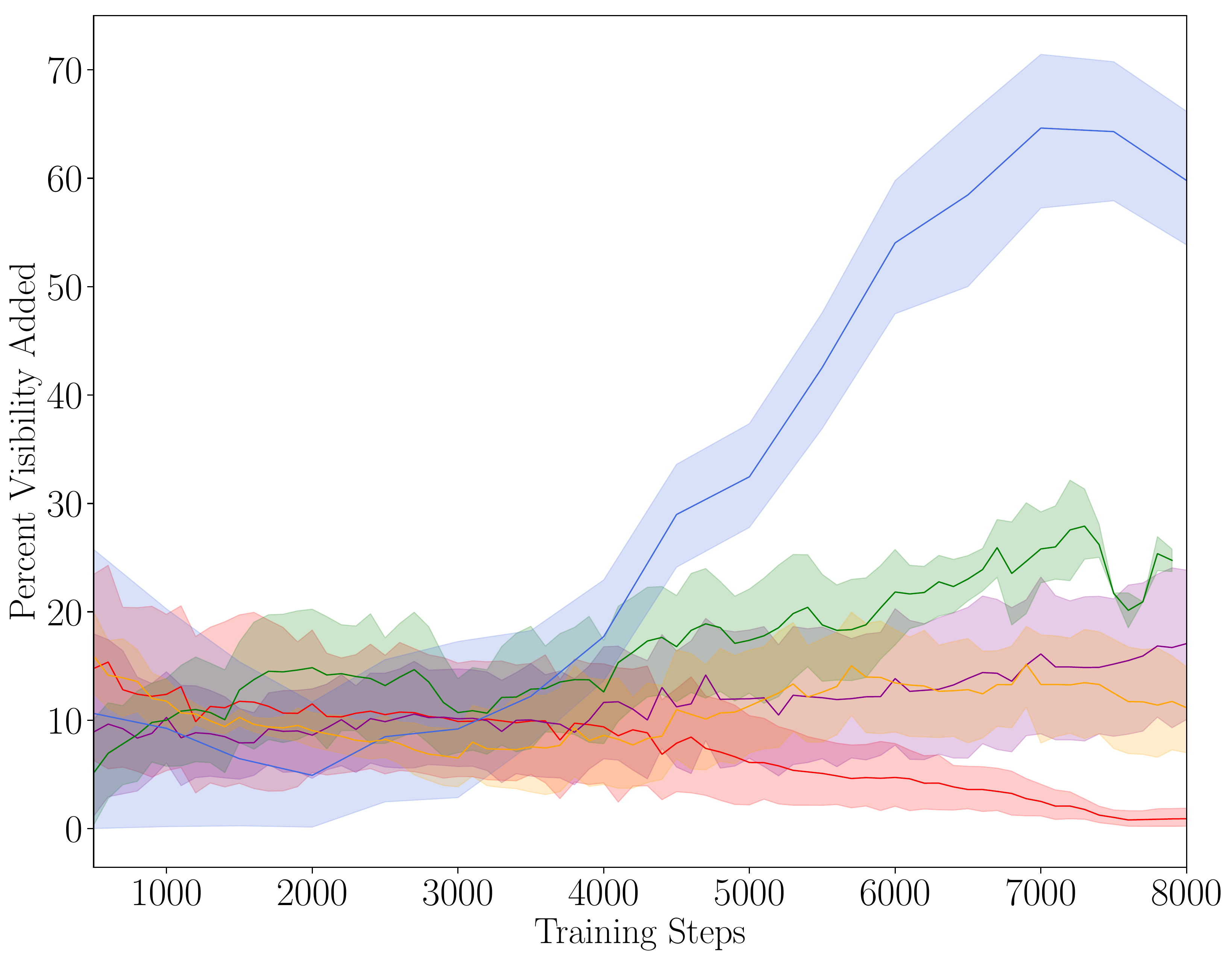}
  \end{subfigure}
  \caption{Ablation results, for (left) single heap condition, and (right) dual heap condition. For the 'No Pose Input' ablation, we provide the agent with an RGB image and the label of the object instead of the approximate target position input. In the single heap setting, the agent can often perform well without the position input by just de-occluding the objects in the heap, but in the two-heap setting this is more important. In both conditions, teacher guidance and asymmetric learning contribute to effective learning.}
  \label{fig:ablation}
  \vspace{-15pt}
\end{figure*}

\subsection{Simulation Experiments} 
\label{sec:sim-exp-procedure}

\paragraph*{Heap Generation} We create two sets of experimental conditions: single-heap and dual-heap. The former has just a single heap of objects near the center of the workspace, while the latter has two heaps at some distance from each other. Intuitively, the latter condition better tests that the policy pays attention to the target object and not just the closest heap of objects.

For the single-heap setting, we programmatically generate  3000 heaps composed of 5, 10, and 15 objects. For the dual-heap setting, we generate 1200 heaps having 5,6,7,8,9, and 10 objects per heap. All heaps are made up of objects from the YCB object set \cite{calli2015benchmarking}.  To generate a heap, the Bullet Physics Engine~\cite{coumans2017bullet} is used to simulate dropping random distinct objects onto the table workspace from a fixed height. After all objects come to rest (their velocity nears zero), the modal (accounting for occlusions) and amodal (disregarding occlusions) segmentation masks of each object are used to check the degree to which they are occluded. The first object with a valid amount of occlusion is made to be the target object of that heap. If no object is least 10\% and no more than 90\% occluded then we discard this heap and do not keep it in the final set. These heaps are kept constant throughout our experiments, so that there is a fair comparison between our approach and baselines. 

\paragraph*{Policy Evaluation} Once heaps are generated, they are used to train and evaluate the deep RL agent. Training is done on either single-heap or dual-heap settings, with random sampling of a heap each episode, and with half of the heaps being held out during training to be used during evaluation. Evaluation is done after every N interaction steps with the environment by running the agent's policy (without additive noise or teacher guidance) on an evaluation heap. The agent is allowed to move its end effector at most 5cm in a given step.  In both training and evaluation, the agent begins each episode in a random position at a minimum distance of 0.2 meters from the target object, and is given at most 50 actions to complete the task. To evaluate whether the trained agent can make objects more graspable and not just more visible, we measure the change in mean grasp quality score as measured by the fully convolutional grasp quality network~\cite{satish2019policy} of the best 10 grasps for the image at the beginning and the end of the episode. We report results using 5 seeds in all conditions.

\subsection{Simulation Results}

As shown in figure \ref{fig:sim-results}, our method is able to converge to high improvements in graspability within only 8,000 action executions in the environment in both the single-heap and the dual heap conditions, with the latter being more challenging. Furthermore, this level of performance corresponds to attaining higher environment rewards, the object being significantly more uncovered, and the object uncovered in less time. Lastly, our method does so with less disturbance to the object heap than any of the teachers, as measured by the mean distance that the non target objects in the heap move.

We compare our method to the baseline DDPG algorithm without teacher guidance or asymmetry but with the same inputs as our agent, and find that it does not learn to improve any of these metrics at all within the same time. We restrict our comparison to just the DDPG algorithm so the baseline RL algorithm is the same as in our method, since it could be implemented using newer state of the art algorithms as well. While results can be expected to be better if using superior algorithms, we report results with DDPG due to ease of implementation and results already being positive. 

As shown in figure \ref{fig:ablation}, the components of asymmetry and teacher guidance are both essential to the method being able to learn this efficiently and effectively. While training with RGB input works on par with the alternative of having the intermediate in the single-heap setting, that is not the case in the dual-heap setting, showing that the intermediate representation of where the target object is matters when there is more uncertainty about it. 

\subsection{Qualitative Results}
As shown in figure \ref{fig:intro-fig}, our policy learns to execute complex continuous control that is different to the behavior of the teachers. The policy learns a behavior to approach the object and ``nudge'' the occluding clutter gently, rather than executing a continuous push like the teachers do. We hypothesize this is an effect of the negative reward for moving the non-target object and the lack of ground truth about object poses used by our teachers with privileged information. The agent learns to uncover the target object moving the occluding objects that occlude it as little as possible. 

\begin{figure}[!b]
\centering
    \begin{subfigure}{0.45\linewidth}
    \centering
    \includegraphics[width=\linewidth]{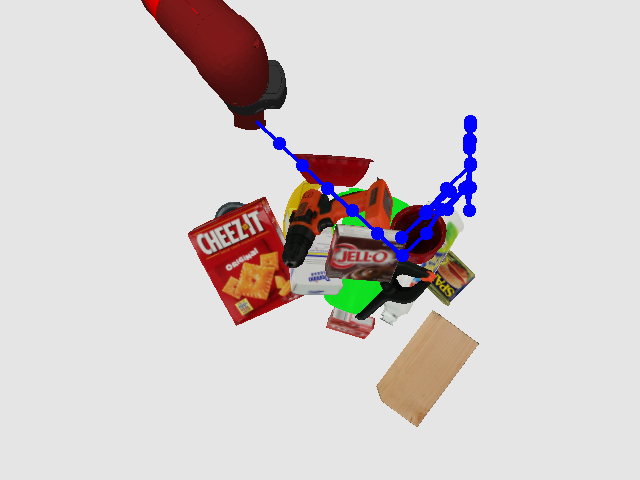}
    \end{subfigure}
  \quad
    \begin{subfigure}{0.45\linewidth}
    \centering
    \includegraphics[width=\linewidth]{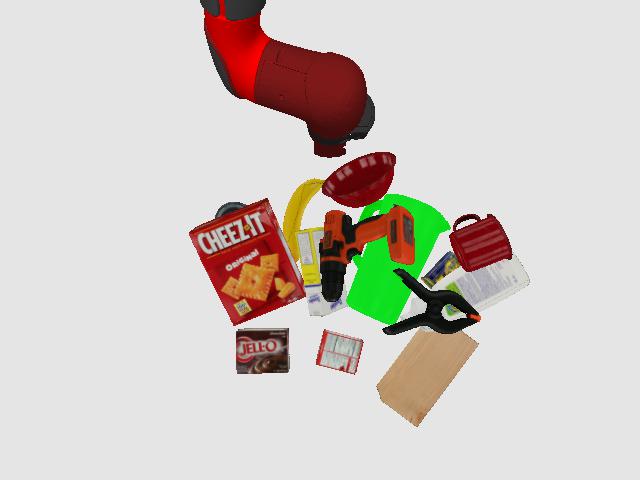}
    \end{subfigure}
  \\ \vspace{0.1cm}
    \begin{subfigure}{0.45\linewidth}
    \centering
    \includegraphics[width=\linewidth]{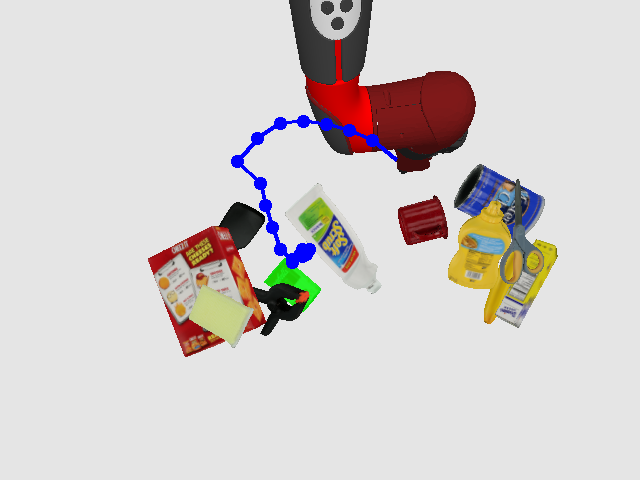}
    \end{subfigure}
  \quad
    \begin{subfigure}{0.45\linewidth}
    \centering
    \includegraphics[width=\linewidth]{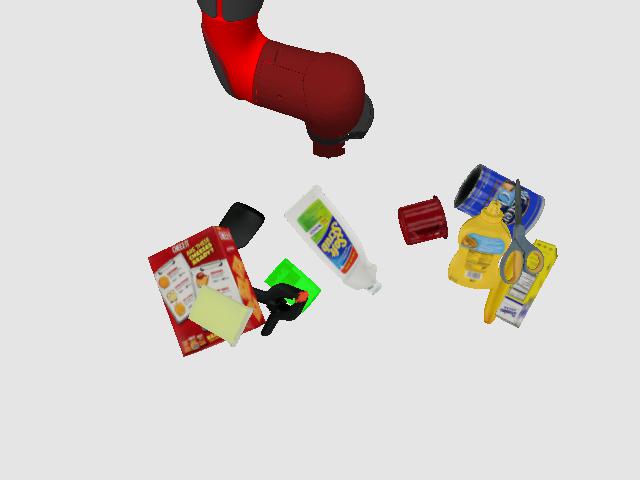}
    \end{subfigure}
  \caption{ 
  Visualizations of failed trajectories from given start states (left) to the end state (right) for two rollouts in simulation. Segmentation mask overlay is shown in green and full trajectory is shown as blue lines on the left column images.}
  \label{fig:qualitative}
  \vspace{-10pt}
\end{figure}

The agents do not achieve full uncovering of the target object and improvement of graspability in all cases due to several failure modes. In some cases, the agent repeats the same actions back-and-forth without causing any change in the environment, as depicted in Fig.~\ref{fig:qualitative}, top row. In other cases the agent moves near the target object but does not interact with the objects occluding it, as shown in Fig.~\ref{fig:qualitative}, bottom row. These failure modes can likely be addressed by modifying the agent's architecture, which we leave to future work.

\section{Conclusion} 
\label{sec:conclusion}
We presented a novel approach combining teacher guidance, asymmetric learning, and a mid-level representation to learn pushing strategies that uncover a known object of interest in cluttered piles of objects. Our learned policies demonstrate behaviors adapted from the teacher demonstrations to the lack of privileged information in the deployment conditions. Our solution achieved positive results, uncovering the target and improving its graspability in most experiment. We plan to work on the failure cases, improving the interaction strategies with better suggested motions from teachers.
\section*{Acknowledgement}
\label{sec:acknowledgement}
\small{
We acknowledge the support of Toyota (1186781-31-UDARO). 
AG is supported in part by CIFAR AI chair. 
We thank our colleagues and collaborators who provided helpful feedback, code, and suggestions, especially Professor Ken Goldberg, Michael Danielczuk, Matt Matl, and Ashwin Balakrishna.
}


\renewcommand*{\bibfont}{\footnotesize}
\printbibliography 

\end{document}